\documentclass[10pt,twocolumn,letterpaper]{article}

\usepackage{wacv}   
\usepackage{times}
\usepackage{epsfig}
\usepackage{graphicx}
\usepackage{amsmath}
\usepackage{amssymb}
\usepackage{booktabs}

\wacvalgorithmstrack   %

\wacvfinalcopy %

\ifwacvfinal
\usepackage[breaklinks=true,bookmarks=false]{hyperref}
\else
\usepackage[pagebackref=true,breaklinks=true,colorlinks,bookmarks=false]{hyperref}
\fi

\usepackage{booktabs}

\usepackage{overpic}
\usepackage{enumitem} %
\usepackage{overpic} %
\usepackage{color}
\usepackage[table]{xcolor}
\usepackage{algorithm,algorithmicx,algpseudocode}
\usepackage{multirow}
\usepackage{comment}
\definecolor{turquoise}{cmyk}{0.65,0,0.1,0.3}
\definecolor{purple}{rgb}{0.65,0,0.65}
\definecolor{dark_green}{rgb}{0, 0.5, 0}
\definecolor{orange}{rgb}{0.8, 0.6, 0.2}
\definecolor{red}{rgb}{0.8, 0.2, 0.2}
\definecolor{darkred}{rgb}{0.6, 0.1, 0.05}
\definecolor{blueish}{rgb}{0.0, 0.3, .6}
\definecolor{light_gray}{rgb}{0.7, 0.7, .7}
\definecolor{pink}{rgb}{0.9, 0, 0.6}
\definecolor{greyblue}{rgb}{0.25, 0.25, 1}
\definecolor{teal}{rgb}{0.0, 0.4, 0.4}

\usepackage{soul}

\usepackage{dsfont}

\newcommand{\Fig}[1]{Fig.~\ref{fig:#1}}
\newcommand{\Figure}[1]{Figure~\ref{fig:#1}}
\newcommand{\Tab}[1]{Tab.~\ref{tab:#1}}
\newcommand{\Table}[1]{Table~\ref{tab:#1}}

\newcommand{\eq}[1]{(\ref{eq:#1})}

\newcommand{\Sec}[1]{Sec.~\ref{sec:#1}}
\newcommand{\Section}[1]{Section~\ref{sec:#1}}

\usepackage{blindtext}

\usepackage{enumitem}
\setlist[itemize]{noitemsep,leftmargin=*,topsep=.05in}
\setlist[enumerate]{noitemsep,leftmargin=*,topsep=.05in}

\renewcommand{\paragraph}[1]{\vspace{.5em}\noindent\textbf{#1}.}

\DeclareMathOperator*{\argmin}{arg\,min}

\newcommand{\loss}[1]{\mathcal{L}_\text{#1}}

\newcommand{\param}{\boldsymbol{\theta}}
\newcommand{\expect}{\mathbb{E}}
\newcommand{\real}{\mathbb{R}}

\newcommand{\kernel}{\mathcal{K}}

\newcommand{\att}{\mathbf{A}}

\newcommand{\point}{\mathbf{p}}
\newcommand{\points}{\mathbf{P}}
\newcommand{\backbone}{\mathcal{F}}
\newcommand{\features}{\mathbf{F}}
\newcommand{\feature}{\mathbf{f}}
\newcommand{\query}{(\point_q, \feature_q)}

\newcommand{\queryFeature}{\feature}

\newcommand{\x}{\mathbf{x}}
\newcommand{\Affinity}{\mathcal{A}}

\newcommand{\cvxnet}{\mathcal{D}}
\newcommand{\pars}{\param}
\newcommand{\cvxFeature}{\mathbf{f}}
\newcommand{\cvxOrigin}{\mathbf{o}}
\newcommand{\cvxNormal}{\mathbf{n}}
\newcommand{\cvxOccupancy}{\mathcal{O}}
\newcommand{\cvxOffset}{d}

\newcommand{\semanticProj}{\phi}
\newcommand{\instanceProj}{\psi}
\newcommand{\semantic}{\mathbf{s}}
\newcommand{\polytope}{\mathcal{C}}
\newcommand{\SDF}{\Phi}

\usepackage{letltxmacro}
\LetLtxMacro{\oldalgorithmic}{\algorithmic}
\LetLtxMacro{\endoldalgorithmic}{\endalgorithmic}
\renewenvironment{algorithmic}[1][0]{%
  \hrulefill\par
  \oldalgorithmic[#1]}
  {\endoldalgorithmic\par
   \vspace*{-.5\baselineskip}
   \hrulefill\par
  }
\begin{document}
\title{NeuralBF: Neural Bilateral Filtering \\ for Top-down Instance Segmentation on Point Clouds}
\author{%
    Weiwei Sun$^{1,3}$\thanks{Work partially done during an internship at Google.} \quad 
    Daniel Rebain$^{1}$\quad
    Renjie Liao$^{1}$ \quad
    Vladimir Tankovich$^{3}$\\
    Soroosh Yazdani$^{3}$ $\quad$
    Kwang Moo Yi$^{1}$\quad
    Andrea Tagliasacchi$^{2,3}$\\[.1in] 
    $^1$University of British Columbia \hspace{3pt}
    $^2$Simon Fraser University \hspace{3pt}
    $^3$Google Research \hspace{3pt}
    \\[.2in]
    \url{https://neuralbf.github.io}
}
\maketitle
\begin{abstract}
We introduce a method for instance proposal generation for 3D point clouds.
Existing techniques typically directly regress proposals in a single feed-forward step, leading to inaccurate estimation.
We show that this serves as a critical bottleneck, and propose a method based on iterative bilateral filtering with learned kernels.
Following the spirit of bilateral filtering, we consider both the deep feature embeddings of each point, as well as their locations in the 3D space.
We show via synthetic experiments that our method brings drastic improvements when generating instance proposals for a given point of interest.
We further validate our method on the challenging ScanNet benchmark, achieving the best instance segmentation performance amongst the sub-category of top-down methods.
\end{abstract}

\section{Introduction}
\label{sec:intro}
Instance segmentation is a critical component of semantic 3D understanding, with applications including robotic manipulation~\cite{he2021deep, xie2021unseen, ogawa2021occlusion, manuelli2019kpam, morrison2018cartman} and autonomous driving~\cite{sirohi2021efficientlps, zhou2020joint, de2017semauto, wong2020identifying, milioto2020lidar}.
An essential step of instance segmentation~\cite{he2017mask, xie2020polarmask, wang2021solo, wang2020solov2, jiang2020pointgroup} is to generate a set of reliable instance proposals.
For natural images, state-of-the-art methods generally follow the \textit{top-down} paradigm~\cite{wang2021solo, wang2020solov2}, where one first \textit{detects} candidate instance proposals and then \textit{prunes} them via non-maximum suppression (NMS). 
Conversely, \textit{bottom-up} methods~\cite{kong2018recurrent, wang2019associatively} learn per-point \textit{embeddings} that are then used to \textit{cluster} points into a disjoint set of proposals. 

It is quite surprising that the dominance of top-down methods in natural images~(2D) is not reaffirmed when we change our domain to point clouds~(3D), where bottom-up methods dominate public leaderboards~\cite{vu2022softgroup, chen2021hierarchical}.
As a consequence, the recent 3D computer vision literature naturally focused on incremental contributions attempting to improve the performance of bottom-up techniques, leaving top-down methods relatively under-investigated.
Therefore, one is left to wonder why such a striking difference in approaching 2D vs 3D instance segmentation exists, and whether it is possible to devise a competitive top-down method.

\begin{figure}%
\centering
\includegraphics[width=\linewidth]{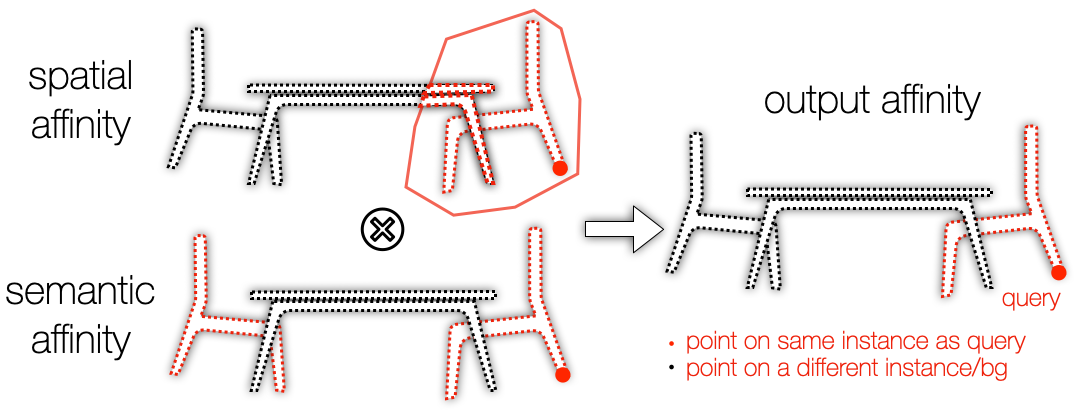}
\\[1em]
\includegraphics[width=.47\linewidth]{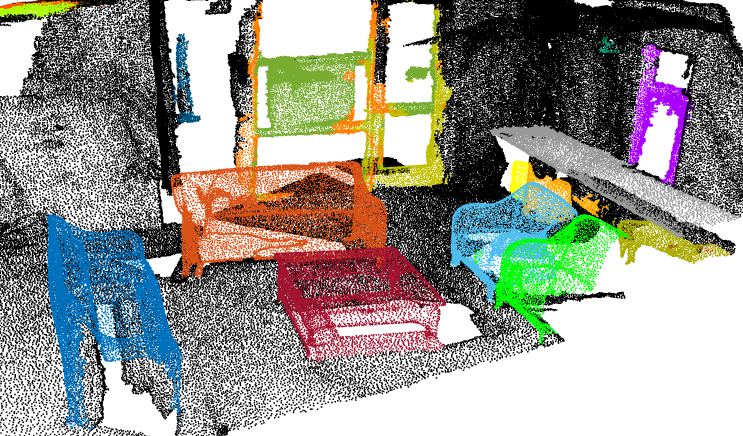}
\hfill
\includegraphics[width=.47\linewidth]{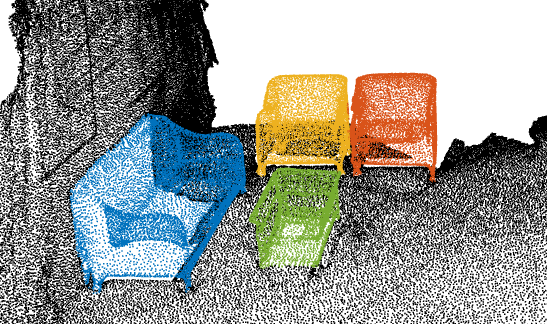}
\vspace{1em}
\caption{
\textbf{Teaser --}
(top) Given a query point, we define the corresponding instance as the element-wise \textit{product} of feature (points with similar class) and spatial (points with similar localization) affinities; this leads to formulating instance segmentation as a neural bilateral filter.
(bottom) Our technique can be applied to large-scale instance segmentation of real-world ScanNet scenes, leading to the best-performance-in-class amongst top-down methods.
}
\label{fig:affinity}
\end{figure}

In this work, we argue that a critical bottleneck exists in the proposal generation process for point clouds.
Early works follow a similar process as for natural images, where bounding boxes are \textit{regressed}~\cite{liu2020learning, hou20193d, yi2019gspn, yang2019learning}, but this regression does not generally lead to sufficiently accurate proposals.
We ablate these techniques on a simple synthetic dataset, demonstrating how they do not lead to satisfactory performance (i.e. mAP${<}50\%$), while we achieve near-perfect results.

Our top-down technique generates the proposal associated to a given \textit{query} on the input point cloud; see~\Figure{affinity}.
We encode a proposal as an \textit{affinity score}: a [0,1] point-wise labeling of the point cloud that is conditional on a query point (i.e. as the query point changes, the affinity scores changes).
We draw inspiration from bottom-up methods~\cite{vu2022softgroup}, and determine two points belong to the same instance if they are ``close'' to each other in \textit{both} space and semantic class; see~\Figure{affinity}.
The \textit{semantic} affinity compares the similarity of semantic features in order to separate points from distinct object types.
The \textit{spatial} affinity is responsible for bounding the spatial extent of the instance in order to separate semantically-similar objects from each other.
Hence, query-conditional affinity can be factorized in two terms, leading us to naturally formulate the problem as a \textit{neural bilateral filter}.

In representing \textit{spatial} affinity, we note the predominant representation employed in the 2D image domain axis-aligned \textit{bounding boxes}.
And while parameterizing spatial affinity with 3D bounding boxes is possible, this either requires careful handling of SE(3) equivariance~\cite{deng2021vector, luo2022equivariant, frameaveraging}, or careful prediction of rotations~\cite{levinson2020rotations}. 
We avoid this issue by introducing the use of differentiable convex hulls~\cite{deng2020cvxnet} for instance proposal.
Note that convex hulls are universal approximators of bounding boxes, capable of implicitly modelling rotated bounding boxes.
From a different standpoint, our technique, which models convexes as the level set of a field, can be viewed as the first method that attempts to apply the rapidly growing area of Neural~Fields~\cite{neuralfields} to 3D instance segmentation.

\paragraph{Contributions}
We validate the effectiveness of our method on both synthetic and real datasets leading to the following contributions:
\begin{itemize}
\item we introduce a simple synthetic dataset that reveals a major bottleneck in instance proposal generation for point clouds;
\item we pose the problem of instance segmentation as generating the affinity of points in the cloud to a query point;
\item we formulate the computation of affinity as a neural bilateral filter, and demonstrate how an iterative formulation improves its performance;
\item we introduce the use of coordinate networks representing convex domains to model the spatial affinity in our neural bilateral filter;
\item collectively, these contributions results in a method that tops the leaderboard in point cloud instance segmentation on ScanNet amongst top-down methods.
\end{itemize}
\section{Related works}
\label{sec:related}
We briefly describe the recent works on 2D and 3D instance segmentation, review methods on mean shift and bilateral kernel, and discuss the recent transformer-based instance segmentation.
For a survey on 3D instance segmentation, please refer to~\cite{he2021deep}, and to \cite{gu2022review} for 2D instance segmentation.

\paragraph{2D instance segmentation}
Top-down methods \cite{he2017mask, xie2020polarmask} predict redundant instance proposals for sampled locations in images, which typically requires NMS to remove the overlap.  
Mask-RCNN~\cite{he2017mask} detects a set of bounding boxes as the initial instance proposals, and then applies a segmentation module and NMS to output the final mask.     
PolarMask~\cite{xie2020polarmask} enhances the performance by using ``center priors'' -- locations close the center of object tends to predict better bounding boxes. 
SOLO~\cite{wang2020solov2, wang2021solo} predict instance masks for every location, obviating the need of segmentation module. 
This is similar to our method where we also output instance masks without segmentation module. 
Other mainstream instance segmentation pipelines~\cite{kong2018recurrent, de2017semantic} follow the bottom-up paradigm clustering pixels into segments as instance proposals, resulting in performance typically inferior to that of top-down methods. 

\paragraph{3D instance segmentation}
In contrast to the 2D image domain, bottom-up methods dominate 3D instance segmentation benchmarks.
PointGroup~\cite{jiang2020pointgroup} first labels points with semantic prediction and center votes, and then cluster points into segments as the instance proposals. 
Follow-up works~\cite{chen2021hierarchical, vu2022softgroup} further enhance the clustering method in different aspects.
HAIS~\cite{chen2021hierarchical} develops hierarchical clustering to have better instance proposals.
SoftGroup~\cite{vu2022softgroup} proposes to group points using soft semantic scores and introduces a hybrid top-down/bottom-up technique via a proposal refinement module.
While bottom-up methods rely on the heuristics such as object sizes and distance threshold, top-down methods largely lag in performance.
Top-down methods \cite{yang2019learning, yi2019gspn} rely on precise bounding box prediction as the initial instance proposal.
In more detail, 3DBoNet~\cite{yang2019learning} directly predicts a fixed set of 3D bounding boxes, while GSPN~\cite{yi2019gspn} proposes a synthesis-and-analysis strategy to predict better bounding boxes.

\begin{figure*}[h]
\centering
\includegraphics[width=\linewidth]{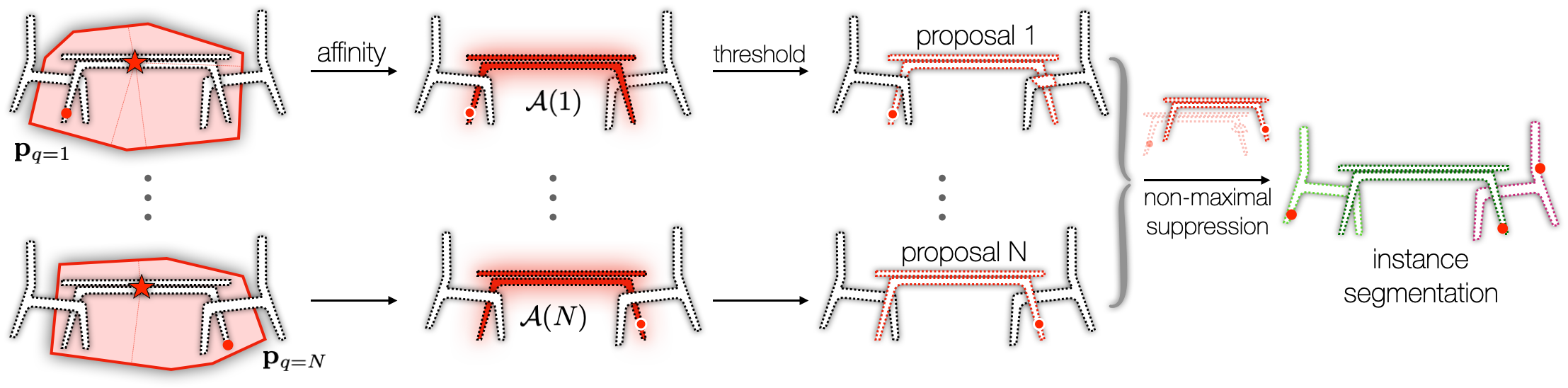}
\caption{
\textbf{Overview -- }
(a) Given a query ({\color{red}$\odot$}), we regress the bounding hull of the corresponding instance;
(b) Together with semantic segmentation, this defines an affinity function on the entire point cloud;
(c) This affinity can be threshold to generate a candidate instance proposal;
(d) Instance proposals are then grouped by non-maximal suppression to generate the scene's instance segmentation. 
}
\label{fig:overview}
\end{figure*}

\paragraph{Neural Bilateral Filtering}
The idea of combining bilateral filtering with neural networks has been mostly in the context of filtering and enhancing natural images~\cite{jampani2016learning, gharbi2017deep, Koshboresh18}.
However, to the best of our knowledge, learned bilateral filtering has not been applied to the context of 3D point clouds, although classical point cloud processing layers for point clouds do exist~\cite{Fleishman03}.

\paragraph{Transformer-based instance segmentation}
The introduction of end-to-end transformer-based instance segmentation~\cite{carion2020end} is revolutionizing the domain of instance segmentation.
For natural images, MaskFormer~\cite{cheng2021per} and Mask2Former~\cite{cheng2022masked} employ transformer architectures to convert queries into a set of unique instance masks, and M2F3D~\cite{m2f3d} applies Mask2Former to 3D point clouds achieving state-of-the-art performance.

\section{Method -- \Figure{overview}}
\label{sec:method}
Given a point cloud of $N$ points in $D$-dimensional space $\points {=} \{\point_n\}$ and corresponding C-dimensional features 
$\features {=}\{\feature_n\} {=} \backbone(\points; \pars_\backbone)$, computed by a deep learning backbone with learnable parameters $\pars_\backbone$, we generate instance proposals by regressing the bounding volume~(i.e. a convex hull in $\real^D$) corresponding to the instance of a query point $(\point_q, \feature_q)$, where $q {\sim} [1,N]$.
Together with segmentation features, bounding volumes imply an affinity~$\att {\in} \real^{N}$ between the query $\query$ and the whole point cloud, which can be thresholded to generate an instance proposal~(\Section{affinity}).
These instance proposals are then aggregated by classical non-maximum suppression (NMS) to generate the desired instance segmentation.

\subsection{Affinity definition}
\label{sec:affinity}
As illustrated in \Figure{affinity}, the affinity of points in the point cloud to a query $\query$ can be intuitively defined as the \textit{element-wise product} of two affinities:
\begin{itemize}
\item \textbf{Affinity in feature space}: whether a point in the point cloud belongs to the same class as the query;
\item \textbf{Affinity in geometric space}: whether a point in the point cloud belongs to the same spatial region as the query.
\end{itemize}
More formally, we define our affinity $\Affinity(q)$ as:
\begin{align}
\Affinity(q) &= 
\Affinity_\point(q) \odot \Affinity_\feature(q),
\label{eq:affinity}
\\
\Affinity_\feature(q)[n] &= 
\exp(-\tau_\feature \cdot \kernel_{\feature}(q,n)),
\label{eq:affinityFeat}
\\
\Affinity_\point(q)[n] &= \exp(-\tau_\point \cdot \kernel_{\point}(q,n)),
\label{eq:affinityPoint}
\end{align}
where $\odot$ is the element-wise product, $[n]$ indexes the $n$-th element of the array, and $\tau$ are hyperparameters controlling the bandwidth of the kernels.
We can then learn the parameters of kernels $\kernel_{\feature}$ and $\kernel_{\point}$, whose internals are provided in what follows, by directly attempting to reproduce the target affinity given a randomly drawn query point:
\begin{equation}
\expect_{q{\sim}[1,N]} 
\left\| \Affinity(q) - \Affinity^{gt}(q) \right\|_1^2
.
\end{equation}

\paragraph{$\kernel_\feature$ -- semantic similarity}
We measure whether two points have similar semantic classes via:
\begin{align}
\kernel_\feature(q,n) = \left\| \semanticProj(\feature_q; \param_\semanticProj) - \semanticProj(\feature_n;
\param_\semanticProj) \right\|_2^2
.
\label{eq:semsim}
\end{align}
where $\semanticProj(\cdot; \param_\semanticProj)$ is a small projection layer with parameters $\param_\semanticProj$ that extracts \textit{semantic similarity features} from the (task agnostic) backbone features $\feature$.

\begin{figure}%
\centering
\includegraphics[width=.8\linewidth]{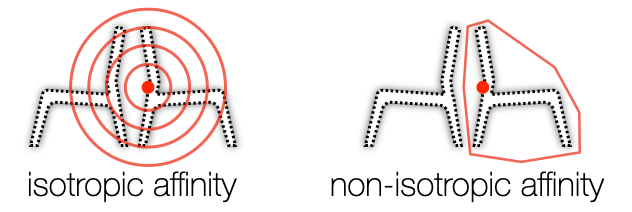}
\caption{
\textbf{Spatial similarity --}
The semantic feature is uninformative in separating the two instances:
(left) an isotropic affinity kernel w.r.t. the query point would mistakenly assign points on the left instance to the right one, regardless of bandwidth choice;
(right) a non-isotropic kernel does not suffer this shortcoming.
}
\label{fig:spatialsimilarity}
\end{figure}
\paragraph{$\kernel_\point$ -- spatial similarity}
While classical bilateral filtering employs \textit{isotropic} kernels to account for spatial similarity~(i.e. gaussian with tunable bandwidth), this is not an optimal choice for 
instance segmentation.
We illustrate our intuition in \Figure{spatialsimilarity}, where the proximity of two objects of the \textit{same} semantic class implies that no isotropic kernel centered at the query point could be used to isolate the desired instance.
We achieve this while retaining \textit{commutative symmetry}\footnote{Affinity ought to be symmetric, because if point $\point_n$ belongs to the same instance as $\point_q$ then we should ideally have $\kernel(q,n)\equiv\kernel(n,q)$.}:
\begin{align}
\kernel_\point(q, n) &= 
\polytope(\point_n{-}\point_q; \instanceProj(\feature_q;\param_\instanceProj))
+ 
\polytope(\point_q{-}\point_n; \instanceProj(\feature_n; \param_\instanceProj))
,
\label{eq:sym_spatial}
\end{align}
where $\instanceProj(\cdot; \param_\instanceProj)$ is a small projection layer with parameters $\param_\instanceProj$ that extracts \textit{spatial similarity features} from the generic backbone features $\feature$.
This leads us to the question of how to design the function~$\polytope(\x; \cvxFeature)$.
One potential solution is to define $\polytope$ as a \textit{coordinate neural network}~\cite{neuralfields} whose shape is described by the feature $\cvxFeature$, and that is evaluated at location $\x$.
We opt to model $\polytope$ with CvxNet~\cite{deng2020cvxnet} -- coordinate neural network architecture that is a universal approximator of convex domains.
This choice is particularly well-suited, because:
\begin{itemize}
\item convex hulls are a topologically equivalent, yet more flexible and detailed replacement for 2D/3D bounding boxes, the core representation employed in 2D/3D object detection/instance segmentation, making them a particularly well-suited choice for our problem;
\item compared to coordinate neural networks implemented as multi-layer perceptrons, CvxNet-like hyper-networks generate very small output networks and are more memory efficient, allowing us to use larger mini-batch sizes leading to faster training.
\end{itemize}
We further detail the design of $\polytope$ in~\Section{convexes}, which will fulfill the following base property with respect to the convex domain specified by the feature~$\cvxFeature$:
\begin{align}
\polytope(\x; \cvxFeature)
\begin{cases}
=0& \text{if } \x \text{ inside convex defined by } \cvxFeature, 
\\
>0& \text{otherwise } (\approx \text{boundary distance).}
\end{cases}
\end{align}

\subsection{Convex parameterization~$\polytope(\x; \feature)$}
\label{sec:convexes}
From $\cvxFeature$, via a fully connected decoder (with \textit{shared} parameters $\pars_\cvxnet$), we derive the normals~$\{\cvxNormal_h {\in} \real^D ~|~ \left\|\cvxNormal_h\right\|_2{=}1\}$ specifying the $H$ half-space orientations, and their distances~$\{\cvxOffset_h {\in} \real^+\}$ from the origin~$\cvxOrigin {\in} \real^D$:
\begin{align}
\cvxOrigin, \{\cvxNormal_h\}, \{\cvxOffset_h\} = \cvxnet(\cvxFeature; \pars_\cvxnet)
,
\label{eq:cvxdec}
\end{align}
and define the distance of $\x$ from the $h-$th hyperplane as:
\begin{align}
\mathcal{H}_h(\x) =
\mathbf{n}_h \cdot ( \x + \cvxOrigin) + \cvxOffset_h
,
\label{eq:hplane}    
\end{align}
which can be assembled into an (approximate, see~\cite{deng2020cvxnet}) distance function from the convex polytope as:
\begin{align}
\SDF(\x; \cvxFeature) =
\max_h\{\mathcal{H}_{h}(\x) \}
,
\label{eq:polytope}
\end{align}
finally leading to our convex spatial proximity:
\begin{align}
\polytope(\x; \cvxFeature)
= 
\max(\SDF(\x; \cvxFeature), 0),
\label{eq:dist}
\end{align}
which can then, if necessary, be converted as an indicator function (i.e. occupancy) for a convex~\cite{deng2020cvxnet}:
\begin{align}
\cvxOccupancy(\x; \cvxFeature) = \text{sigmoid}(-\SDF(\x; \cvxFeature))
.
\label{eq:occupancy}
\end{align}

\begin{figure}[t]
\begin{algorithmic}[1]
\State \textbf{Input}: \\
\quad $q \in [1,N]$ \Comment{query index} \\
\quad $\points \in \real^{N\times D}$ \Comment{(const) cloud positions} \\
\quad $\features \in \real^{N\times C}$ \Comment{(const) cloud features}
\Function{NeuralBilateralFilter}{}
\vspace{+0.2em}
\State $\point^{(0)} = \point_q = \points[q]$
\State $\feature^{(0)} = \feature_q = \features[q]$ 
\For{$t=1,\ldots, T$} 
    \State $\att^{(t-1)}(q) = \Affinity(\point^{(t-1)}, \feature^{(t-1)})$
    \State $\att^{(t-1)}(q) = \att^{(t-1)}(q) / \|\att^{(t-1)}(q)\|_1$
    \State $\point^{(t)} = \att^{(t-1)}(q) \cdot \points$
    \State $\feature^{(t)} = \att^{(t-1)}(q) \cdot \features$
\EndFor
\State \textbf{Return}  $\att^{(T)}(q)$
\EndFunction 
\end{algorithmic}
\caption{
\textbf{Neural Bilateral Filter --}
Given a query, consisting of a position and corresponding feature, we iteratively apply the learned filters to advect the query point position and features.
Ultimately, the final attention $\att^{(T)}(q)$ is used for downstream tasks.
}
\label{fig:nbf}
\end{figure}

\subsection{Neural Bilateral Filter -- \Figure{nbf}}
\label{sec:bilateral}
The resemblance of~\eq{affinity} to the product of kernels in \textit{bilateral filtering}~\cite{he2012guided,tomasi1998bilateral} inspired us to investigate the use of \textit{iterative} inference.
Specifically, given a query, we advect \textit{both} query position and features, where the advection weights are given by the affinity definition from \eq{affinity}.
Note that the point cloud $\points$ and corresponding features $\features$ remain \textit{unchanged}, only the query is affected.
The outcome is simply that, rather than attention $\att^{(0)}(q){=}\Affinity(q)$ in downstream processing, $\att^{(T)}(q)$ will instead be used.

\subsection{Training}
\label{sec:supervision}
\noindent
To train our network, we optimize: 
\begin{equation}
\argmin_{\pars_\semanticProj, \pars_\instanceProj, \pars_\cvxnet, \pars_\backbone} \quad \loss{affinity} + \loss{sem} + \loss{poly} + \loss{shift}
.
\end{equation}
Of these losses $\loss{affinity}$ is our core loss, while the rest provide ``skip-connection'' supervision to the network to facilitate learning.
Since our method performs iterative inference, we adopt the strategy from ~RAFT~\cite{teed2020raft}, and \textit{discount}~($\alpha{=}0.8$) contributions of later iterations so to encourage faster convergence:
\begin{align}
\loss{affinity} = \expect_{q{\sim}[1,N]} \sum_{t=1}^{T}
\alpha^t \left\| \att^{(t)}(q) - \att^{gt}(q) \right\|_1^2
.
\label{eq:loss_affinity}
\end{align}
\begin{figure}[t]
\centering
\includegraphics[width=\linewidth]{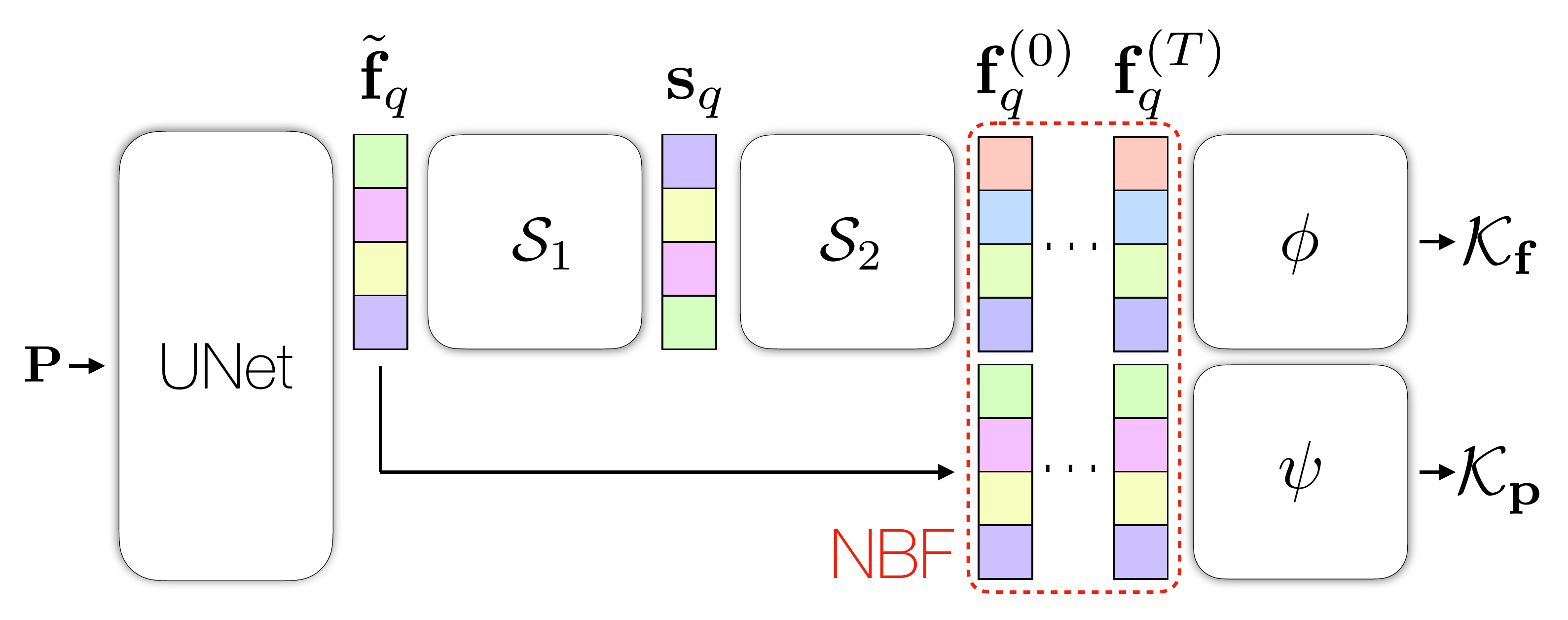}
\vspace{-1em}
\caption{
\textbf{Architecture --}
The point cloud is processed by a backbone to produce $\feature_q{=}\feature_q^{(0)}$, which is then processed by a neural bilateral filter by our kernels $\kernel_*$.
We supervise $\mathbf{s}_q$ via ground truth semantic classification labels, where $\mathcal{S}_1$ is a 2-layer MLP, and $\mathcal{S}_2$ is a linear layer.
}
\label{fig:architecture}
\end{figure}

\paragraph{Semantic supervision}
To encourage the semantic features in \eq{semsim} to represent \textit{only} the semantic similarity, we inject semantic information by mapping intermediate backbone feature to semantic logits (see \Figure{architecture}), and supervising with ground truth labels $\semantic_q^{gt}$:
\begin{align}
\loss{sem} = \expect_{q{\sim}[1,N]}
\left[
\text{CrossEntropy}(\semantic_q, \semantic_q^{gt})
\right]
.
\label{eq:loss_sem}
\end{align}

\paragraph{Instance centroid supervision}
To minimize the learning complexity of $\cvxnet$, we incentivize predicted convex hulls to be expressed with respect to a \textit{stable} coordinate frame\footnote{If two points $a$ and $b$ belong to the same instance, then the predicted convex origin $\cvxOrigin_a \equiv \cvxOrigin_b$, and the same half-space configuration can be used for all queries within an instance; Note this is similar to the coordinate frame normalization in NASA~\cite{nasa}.}.
We employ the ground truth instance origin $\mathbf{c}_q^{gt}$ and supervise the predicted origin relative offset:
\begin{align}
\loss{shift} =
\expect_{q{\sim}[1,N]}
\sum_{t=1}^T \alpha^t
\left\|
(\point_q + \cvxOrigin_q^{(t)}) - \mathbf{c}_q^{gt}
\right\|_1
,
\end{align}
where the offset $\cvxOrigin_q^{(t)}$ is computed from $\cvxnet(\cvxFeature_q^{(t)})$.

\paragraph{Convex occupancy supervision}
Note the affinity supervision in~\eq{loss_affinity} only penalizes points that are incorrectly marked as \textit{outside} the convex hull.
To correct this, let~$\cvxOccupancy_q^\text{gt}(p)$ be the ground truth occupancy of point $p$ belonging to the convex hull of query $q$, we then penalize:
\begin{align}
\loss{poly} = \expect_{q} \expect_{n} \sum_{t=1}^T
\omega_{q,n}
\alpha^{t}
\|
\cvxOccupancy(\point_n{-}\point_q; \instanceProj(\feature_q^{(t)}))
{-} 
\cvxOccupancy_q^\text{gt}(p)
\|_2^2
,
\end{align}
where $\omega_{q,n}$ is a term to control class imbalance: if the instance corresponding to $q$ has $Q$ points and the scene has $N$ points, then $\omega_{q,n}{=}1/Q$ if point $n$ belongs to the instance, and $\omega_{q,n}{=}1/(N{-}Q)$ otherwise.

\subsection{Implementation details}
\label{sec:details}
\noindent
We briefly discuss the core implementation details. A public implementation of our method is available at: \\  \url{https://anonymized.link}.

\paragraph{Network architecture} 
For the backbone we utilize the U-Net-like backbone in \cite{jiang2020pointgroup, chen2021hierarchical} which is implemented with sparse convolution \cite{graham20183d}. 
The dimension $C$ of the backbone feature $\feature$ is 32.

\paragraph{The projection layers $\semanticProj(\cdot;\pars_\semanticProj)$ in \eq{semsim} and $\instanceProj(\cdot;\pars_\instanceProj)$ in \eq{sym_spatial}}
The layer $\semanticProj$ is composed of semantic layers ($\mathcal{S}_1$) and an embedding layer ($\mathcal{S}_2$).
The semantic layers convert the backbone features into  semantic scores with a two-layer MLP with 32 neurons and then outputs the semantic feature with a linear layer of $C$ neurons. 
Note that during the iterative process, we directly update the query's semantic feature without re-using the semantic branch. 
The embedding layer is a linear layer of $C$ neurons.
$\instanceProj(\cdot;\pars_\instanceProj)$ is also a single linear layer with $C$ neurons.

\paragraph{The polytope network $\cvxnet(\cdot;\pars_\cvxnet)$ in \eq{sym_spatial}}
The network $\cvxnet$ consists of two MLP blocks.
The first block -- a two-layer ReLU-activated MLP with 128 neurons -- predicts $\mathbf{o}$ from the query feature $\queryFeature$ and a residual to $\queryFeature$.
We then add the residual to $\queryFeature$ and utilize the second block -- a three-layer ReLU-activated MLP with 128 neurons -- to predict normals and offsets.
For predicting the plane offset $d_h$, we use the strategy from~\cite{fu2018deep} and discretize the offset values into $32$ equal bins in the range $[0, 8]$ meters, and obtain the predicted value via the weighted sum of classification scores.
We represent each 3D convex polytope with twelve planes, striking a good balance between precision and computational load, which linearly increases with the number of planes.
Finally, $\tau_\features{=}1$ and $\tau_\points{=}50$ in \eq{affinityFeat} and \eq{affinityPoint}.

\paragraph{Forming the training batch}
While possible, training with all points in the point cloud is impractical and inefficient, as it would create a quadratic increase in both memory and computation.
We use a batch of four scenes, and randomly sample 32 random points/scene during training to form a single training sample.
We further set the number of mean shift iterations $T{=}2$, which we ablate in~\Sec{ablation}.

\paragraph{Training}
As in RAFT~\cite{teed2020raft}, we detach the gradient flow between different iterations to stabilize training.
We use the Adam optimizer~\cite{kingma2014adam} with cosine annealing for the learning rate \cite{loshchilov2016sgdr}, with 0.001 as the initial learning rate.
We further follow standard data augmentation/voxelization schemes of existing instance segmentation methods~\cite{chen2021hierarchical}.

\paragraph{Non-maximum suppression}
To obtain the final instance segmentation results for the ScanNet dataset we use standard non-maximum suppression~\cite{wang2021solo, ren2015faster} to remove redundant proposals.
In more details, we visit a queue of input candidate proposals in confidence score order; see~\Sec{scannet}.
For each candidate proposal, we compute the IoU with all other candidates and merge/prune those that have IoU higher than 0.25.
\begin{table*}[t]
\centering
\resizebox{.95\linewidth}{!}{ %
\begin{tabular}{lccccccccc}
\toprule
                           & \multicolumn{3}{c}{\textit{Line segment}} & \multicolumn{3}{c}{\textit{Circle}} & \multicolumn{3}{c}{\textit{Average}} \\
                               \cmidrule(lr){2-4} \cmidrule(lr){5-7} \cmidrule(lr){8-10} 
                           & mAP    & AP$_{50}$   & AP$_{25}$   & mAP     & AP$_{50}$& AP$_{25}$   & mAP     & AP$_{50}$ & AP$_{25}$    \\
\midrule
BBox & $46.4_{\pm 1.1}$   & $67.7_{\pm 1.9}$   & $69.8_{\pm 1.1}$     & $21.2_{\pm 1.4}$    & $54.7_{\pm 2.3}$    & $90.6_{\pm 0.3}$   & $33.8_{\pm 0.9}$    & $61.2_{\pm 1.7}$    & $80.2_{\pm 0.7}$    \\
BBox w/ center & $54.1_{\pm 1.6}$  & $77.9_{\pm 1.5}$   & $80.4_{\pm 1.2}$     & $28.0_{\pm 0.8}$    & $64.0_{\pm 0.7}$    & $89.2_{\pm 0.7}$   & $41.0_{\pm 1.0}$    & $71.0_{\pm 0.7}$    & $84.8_{\pm 0.8}$    \\
BBox + GT filtering & $53.9_{\pm 1.4}$   & $68.2_{\pm 1.6}$   & $69.0_{\pm 1.1}$     & $31.9_{\pm 1.9}$    & $71.1_{\pm 1.7}$    & $91.7_{\pm 0.5}$   & $42.9_{\pm 1.2}$    & $69.7_{\pm 1.2}$    & $80.3_{\pm 0.5}$    \\
BBox w/ center + GT filtering & $65.3_{\pm 1.7}$   & $79.3_{\pm 1.5}$   & $80.1_{\pm 1.5}$     & $41.4_{\pm 1.1}$   & $75.4_{\pm 1.4}$    & $90.3_{\pm 0.5}$   & $53.3_{\pm 1.1}$    & $77.3_{\pm 0.6}$    & $85.2_{\pm 0.9}$    \\
Ours          & $\textbf{95.9}_{\pm 0.3}$   & $\textbf{97.6}_{\pm 0.4}$   & $\textbf{97.9}_{\pm 0.3}$   & $\textbf{98.2}_{\pm 0.5}$    & $\textbf{98.9}_{\pm 0.3}$      & $\textbf{99.3}_{\pm 0.3}$   & $\textbf{97.1}_{\pm 0.2}$    & $\textbf{98.3}_{\pm 0.2}$    & $\textbf{98.6}_{\pm 0.1}$   
\\
\bottomrule
\end{tabular}
}
\caption{
\textbf{Query-conditioned instance proposal generation} --  
we randomly sample a single query for each instance and generate the non-overlapped proposals.  
We report the mean and standard deviation of average precision by running the evaluation pipeline five times.
} %
\label{tab:2d}
\end{table*}
\section{Results}
\noindent
In our results section, we:
\begin{itemize}
\item \Sec{2d} -- validate our method in a controlled \textit{synthetic} setup, where we show that current proposal generation methods have limited effectiveness;
\item \Sec{scannet} -- demonstrate the potential of our method in a more complex instance segmentation pipeline on the \textit{real-world} ScanNet dataset~\cite{dai2017scannet}; 
\item \Sec{ablation} -- perform an \textit{ablation} study.
\end{itemize}

\subsection{Synthetic dataset}
\label{sec:2d}
\begin{figure}
\begin{center}
\includegraphics[width=.98\columnwidth]{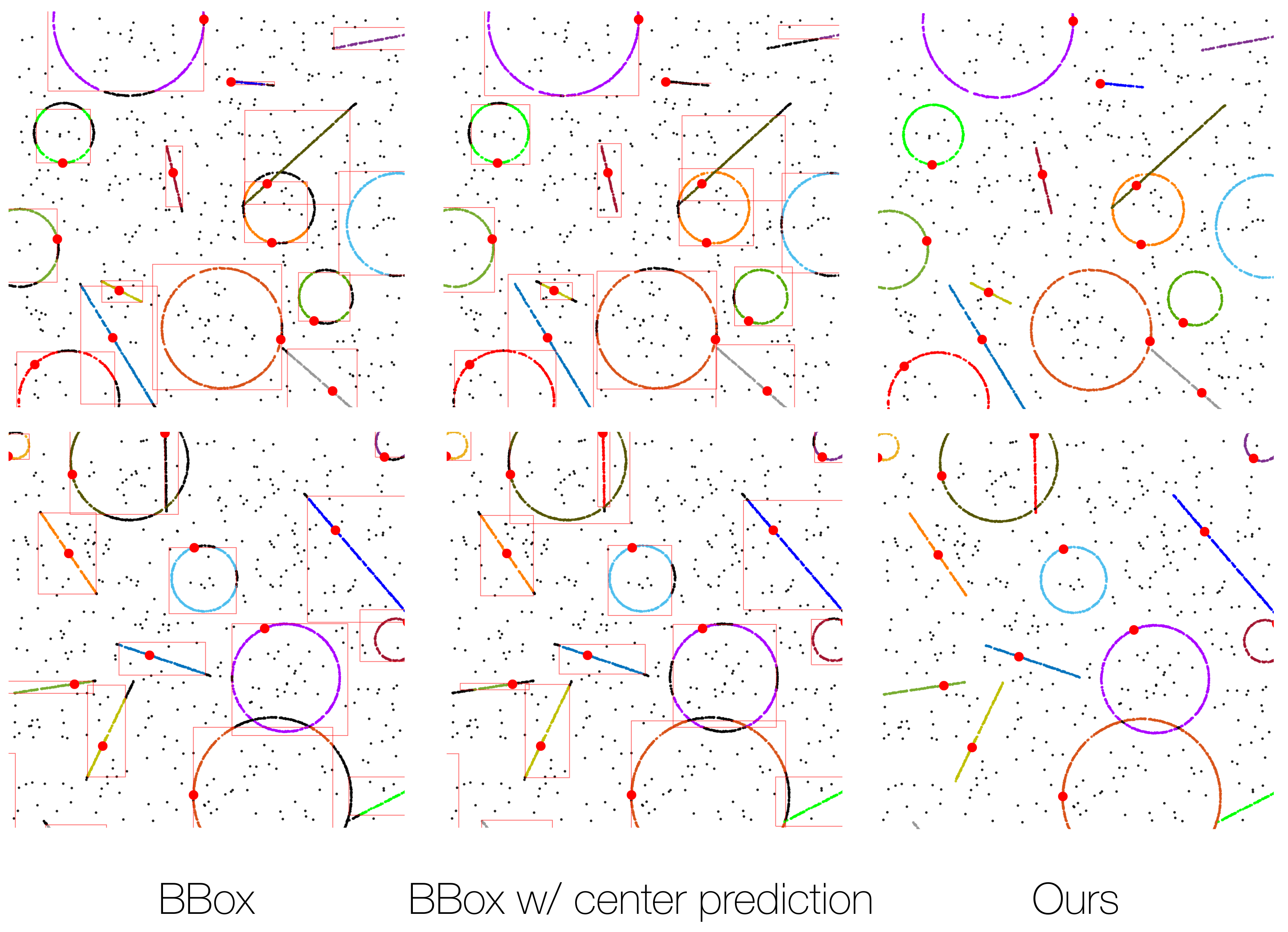}
\end{center}
\vspace{-1em}
\caption{
\textbf{Qualitative/Synthetic -- }
Our algorithm generates the nearly perfect query-conditioned instance proposals while the baseline is limited by the noisy bounding box.    
Note the red large points are the sampled queries. We color the points detected by different instance proposals. 
And the black points are the background points or the points detected by more than two instance proposals. 
}
\label{fig:2d_res}
\end{figure}

We create a 2D synthetic dataset composed of lines, circles, and random noise; see~\Fig{2d_res}.
For each scene, we randomly place $16$ primitives sampled from a large pool~(10k in total) of randomly generated line segments and circles in a 2D space. 
We sample 4096 points for foreground instances and 512 points for the background noise.
To keep a similar point density for instances of different sizes, we make the number of points for each instance proportional to the length of the primitive instance. 
We generate these scenes on-the-fly while training and keep $100$ scenes for testing.
We limit the 2D coordinates to be within $[-4, 4]$ to match the typical size of ScanNet scenes, allowing us to reuse the same backbone across both synthetic and real scenes.

\paragraph{Metrics}
With the dataset, to show that instance proposals are a bottleneck, we are interested in their \textit{direct} evaluation without any downstream Non-Maximum Suppression~(NMS) heuristic.
We randomly select a \emph{single} point for each instance in the point cloud and measure the quality of the generated proposal for the selected point.
Once the proposals are provided, we use the standard metrics used in the ScanNet benchmark~\cite{dai2017scannet}---AP$_{50}$ and AP$_{25}$, which  are the accuracy computed with the intersection-over-union (IoU) threshold of $50\%$ and $25\%$, respectively, and mAP, which is the average AP over different thresholds ranging from $50\%$ to $95\%$ with the step size of $5\%$.

\paragraph{Baselines}
A commonly-used baseline is to directly predict the bounding box for each instance, within which post-processing is applied~\cite{liu2020learning,hou20193d,yi2019gspn,yang2019learning}.
To do so, similarly to GICN~\cite{liu2020learning}, we train a 2-layer MLP that predicts the bounding box, parameterized by its two corners relative to the query.
We further compare against VoteNet~\cite{qi2019deep}, where one first regresses a spatial offset given a query point and then regresses the bounding box corners relative to the offset.
\quad
For these baselines, the bounding box often contains points from noise or other classes (lines vs circle), so we utilize the semantic predictions from the backbone to filter those points out of each instance proposal.
Clearly, this would not perfectly filter out cases where the same class instances overlap; hence, we further propose an oracle baseline which uses \textit{ground-truth} semantic and instance labels as an oracle for filtering, thus emulating an ideal post-processing step for the methods based on bounding boxes.
We train with 10k iterations, which is enough for all methods to converge on this simple dataset.

\paragraph{Results -- \Tab{2d} and \Fig{2d_res}}
Our method outperforms the baselines by a significant margin.
Despite the success of bounding box proposals in 2D images, these method achieves a surprisingly low performance on this simple synthetic dataset,
even when ground-truth filtering is employed.
On the other hand, our method delivers near-perfect results, as one would expect for such a simple dataset.
\quad
For the baselines, as shown in the examples in \Fig{2d_res}, we find that many of the proposals are slightly off, with some being \textit{completely} off.
While small errors in bounding box position/size are not critical for detection in 2D images, they can be catastrophic for point clouds sampled from the object surface near the bounding box exterior, where a small misalignment could remove entire sections of geometry.
For example, in \Fig{2d_res} top-left, the bottom right circle is detected with a bounding box that would be considered quite accurate should one consider only the bounding box, but the majority of the point cloud points for this circle lie outside of the bounding box, as the box is slightly smaller than the actual circle.

\begin{figure}
\begin{center}
\includegraphics[width=\linewidth]{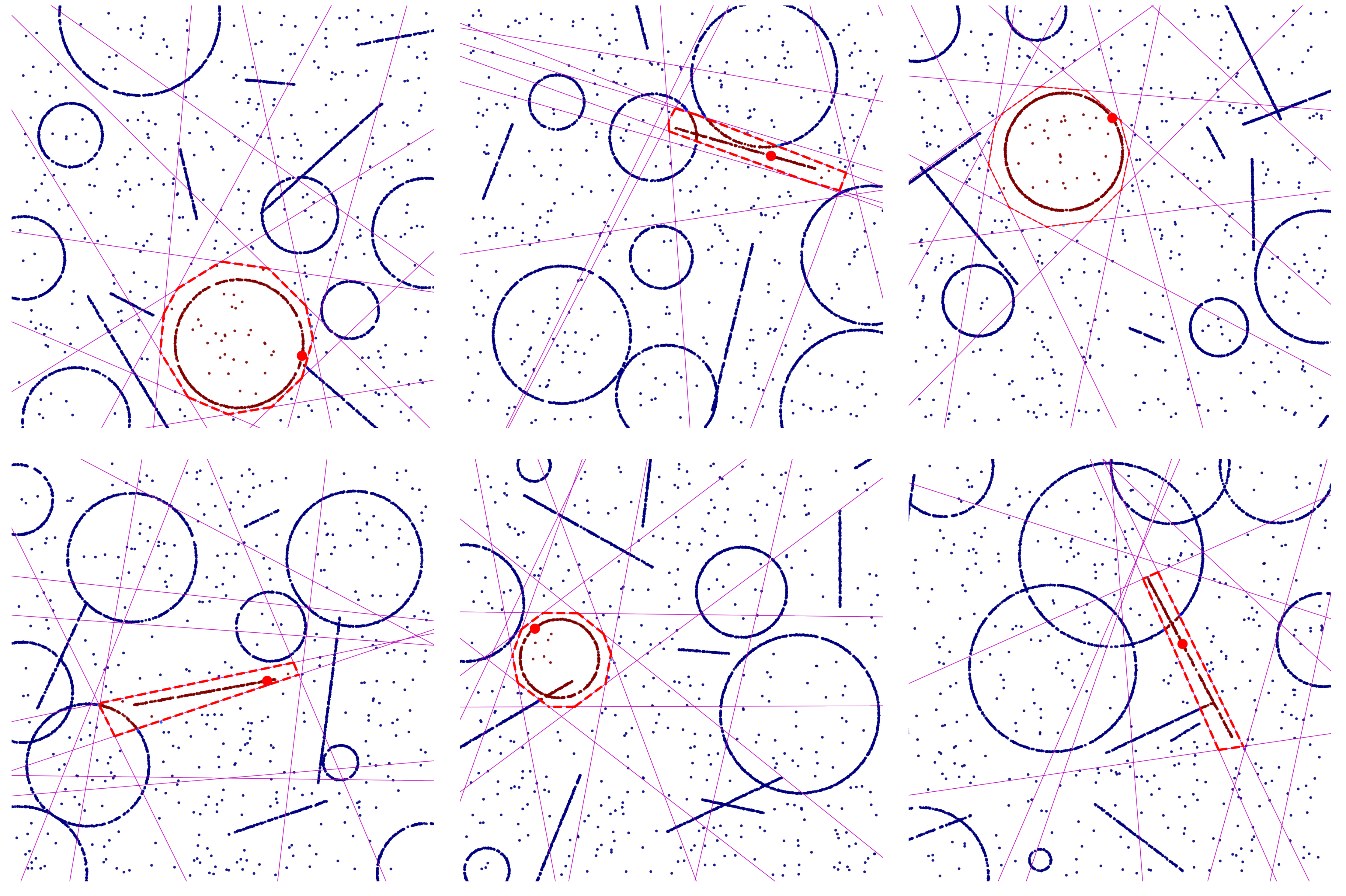}
\end{center}
\vspace{-1em}
\caption{
\textbf{Visualizing the spatial kernels -- } 
Our method learns the effective convex hulls that act as the tight bounding box of the target instance. 
For each convex hull, the magenta lines are the learned half-planes. The red polygon is the intersection between half-planes.  
Points are colored with spatial similarity where red means larger similarity while blue means smaller similarity.
}
\label{fig:cvx_hulls}
\end{figure}
\paragraph{Visualizing the spatial kernels -- \Fig{cvx_hulls}}
We visualize the learned spatial kernel.
As shown, the learned spatial kernel forms a polytope that tightly bounds the instance in question as desired.
These learned kernels enable our method to easily separate different instances spatially, even without considering semantics. 
Such easy separation would not be possible, for example with standard Euclidean distance as points far away from each other on the line or on the circle would be confused with other nearby points.

\begin{table}[h]
\resizebox{\linewidth}{!}{ %
\begin{tabular}{llcccccc}
\toprule
            &Methods & \multicolumn{3}{c}{Validation} & \multicolumn{3}{c}{Test} \\
            \cmidrule(lr){3-5} \cmidrule(lr){6-8}
        && mAP    & AP$_{50}$   & AP$_{25}$   & mAP     & AP$_{50}$& AP$_{25}$   
        \\
\midrule
\multirow{3}{*}{Bottom-up}
& PointGroup~\cite{jiang2020pointgroup} & $34.8$ & $56.7$               & $71.3$               & $40.7$               & $63.6$               & $77.8$               \\
& SSTNet~\cite{liang2021instance}     & $49.4$ & $64.3$               & $74.0$               & $50.6$               & $69.8$               & $78.9$               \\
& HAIS~\cite{chen2021hierarchical}         & $43.5$  & $64.4$               & $75.6$               & $45.7$               & $69.9$               & $80.3$               \\
\midrule
\multirow{2}{*}{Mix} 
& Dyco3D~\cite{he2021dyco3d}     &$35.4$  & $57.6$               & $72.9$               & $39.5$               & $64.1$               & $76.1$               \\
& SoftGroup~\cite{vu2022softgroup}  &$-$  & $67.6$               & $78.9$               & $50.4$               & $76.1$               & 86.5   \\

\midrule
\multirow{4}{*}{\textbf{Top-down}}
& 3D-SIS~\cite{hou20193d}     & $-$  & $18.7$               & $35.7$               & $16.1$               & $38.2$               & $55.8$               \\
& GSPN~\cite{yi2019gspn}       &$19.3$  & $37.8$               & $53.4$               & $-$                  & $30.6$               & $-$                  \\
& 3D-Bonet~\cite{yang2019learning}   &  $-$& $-$                  & $-$                  & $25.3$               & $48.8$               & $68.7$               \\
& \textbf{Ours}   &\textbf{36.0}  & \textbf{55.5}                  & \textbf{71.1}                  & \textbf{35.3}               &  \textbf{55.5}              & \textbf{71.8}               \\
\bottomrule
\end{tabular}}
\caption{
\textbf{Quantitative/ScanNetV2 --} instance segmentation benchmark; our method provides the top-performing solution for the top-down category. For looser thresholds our method performs slightly worse, which may be improved with more sophisticated post-processing.
} %
\label{tab:scannet}
\end{table}

\subsection{Instance segmentation on ScanNetV2}
\label{sec:scannet}
The ScanNetV2~\cite{dai2017scannet} dataset consists of $1613$ scenes in total with $1201$, $312$, and $100$ scenes dedicated for training, validation, and testing, respectively. 
We use the standard evaluation pipeline and report the standard metrics, the same ones as the ones used for the 2D synthetic data.
To evaluate our method for \textit{top-down} instance segmentation pipeline for point clouds, we introduce basic post-processing steps that are commonly used in the literature~\cite{tian2019fcos, wang2021solo, wang2020solov2, liu2020learning}, as well as a scoring function to provide confidence scores for each instance proposal, as required by the benchmark protocol.
Notably, our post-processing steps are relatively simple compared to tricks like ``matrix NMS''~\cite{wang2020solov2} and query sampling using ``center priors''~\cite{liu2020learning,tian2019fcos}.
Specifically, we first segment out all background points using  $\semanticProj(\feature_i)$ in \eq{semsim}.
We then sample $256$ query points from the predicted foreground points and generate $256$ instance proposals. 
When sampling queries, we apply farthest point sampling~\cite{qi2017pointnet++} to ensure maximum coverage.
We then remove redundant instance proposals by applying Non-Maximum Suppression (NMS) to instance proposals with an IoU threshold of 30\%.
We train the entire pipeline end-to-end for $500$ epochs as in~\cite{chen2021hierarchical}.

\paragraph{Confidence scores}
As the benchmark protocol requires instance proposals to have associated confidence scores, we provide a confidence score for each proposal based on both the semantic segmentation score (provided by $\semantic_q$) and an MLP that is trained to regress the IoU of each proposal with respect to ground-truth.
Specifically, we train a two-layer MLP with an $\ell_1$ loss for the IoU.
The final confidence score for each proposal is computed by multiplying the regressed IoU value and the average semantic segmentation confidence.

\paragraph{Dropping low-confidence proposals}
In addition to the above, we drop proposals that have low confidence values (i.e. proposals with semantic confidence lower than 0.1, or with estimated IoU less than 0.2).
Furthermore, we drop proposals that have different predicted labels for the proposal and the query point.
These proposals are from points that are often located where two different instances of different classes meet, and hence are unreliable.

\begin{figure}[h]
\begin{center}
\includegraphics[width=.95\columnwidth]{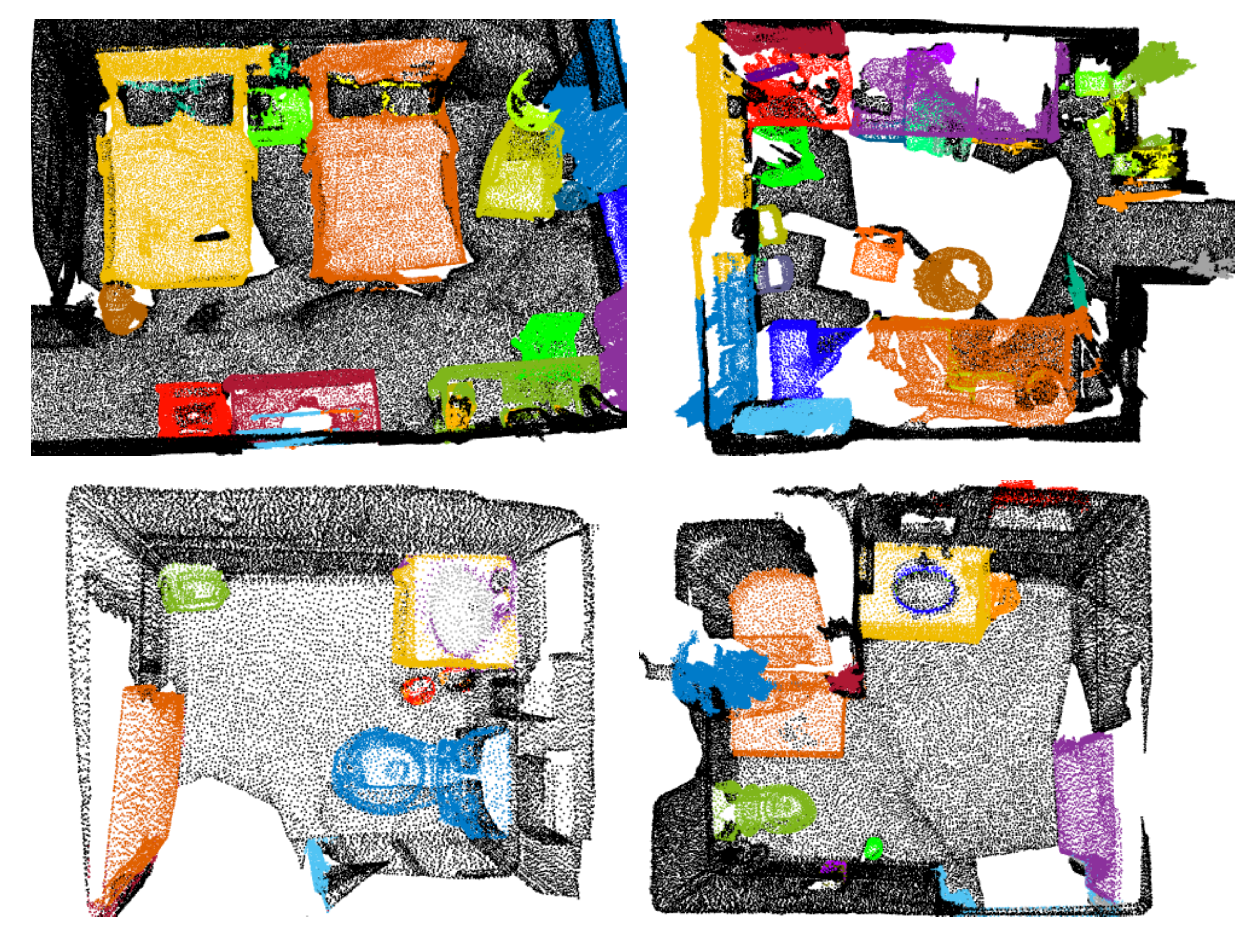}
\end{center}
\vspace{-1em}
\caption{
\textbf{Qualitative/ScanNet -- }
Instance segmentation results on test set.
}
\label{fig:scannet_quali}
\end{figure}
\paragraph{State-of-the-art comparisons -- \Table{scannet} and \Fig{scannet_quali}}
Our method shows promising results compared to the state-of-the-art. 
Among purely top-down methods, our method achieves top performance
validating the effectiveness of our instance proposals generation.
We leave further improvement via better post-processing steps for future work.
\quad
While our method performs worse than the most recent bottom-up methods or hybrid methods~\cite{vu2022softgroup}, we note that these are methods heavily fine-tuned to achieve SOTA benchmark results, whereas ours is not.
Note that our top-down method beats the leading bottom-up method that was SOTA around CVPR 2020~\cite{jiang2020pointgroup}.
Considering that top-down methods have intriguing properties (\eg, their dominant performance in image benchmarks~\cite{wang2021solo,wang2020solov2} and better generalization ability), and are worthwhile to explore further, we believe our work provides progress for instance segmentation on point clouds. 

\begin{figure}
\begin{center}
\resizebox{\linewidth}{!}{ %
\begin{tabular}{lcccc}
\toprule
num\_iter & 1  & 2    & 3    & 4 \\
\midrule
mAP       & $94.46_{\pm 0.55}$ & $\textbf{96.80}_{\pm 0.35}$ & $96.54_{\pm 0.40}$ & $96.18_{\pm 0.68}$
\\  
\bottomrule
\end{tabular}}
\includegraphics[width=\linewidth,trim=0 0 0 -4em]{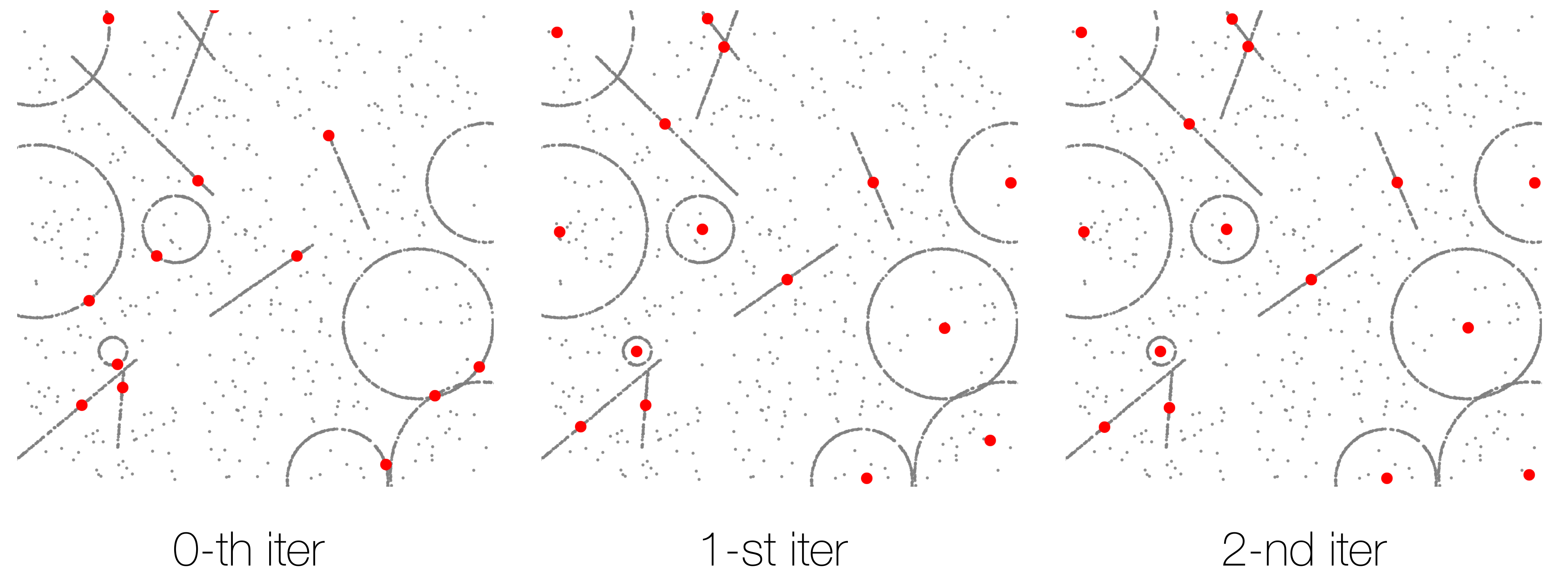}
\end{center}
\vspace{-1em}
\caption{
\textbf{Ablation: number of iterations $T$ --}
(top) We report the average and standard deviation of mAP by repeating the experiment 5 times.
(bottom) Our algorithm shifts the queries (red points) to the centroid after a small number of iterations.
} %
\label{fig:abs_num_iter}
\end{figure}
\subsection{Ablations}
\label{sec:ablation}
\paragraph{Number of iterations -- \Fig{abs_num_iter}}
Our algorithm is capable to shift query points to the center of each instance in just two iterations.
This leads to queries from the same instance to share a similar coordinate frame, leading to a reduction of representation complexity as noted in NASA~\cite{nasa}.
This is beneficial, as a smaller number of iterations reduces the GPU memory load of training.
Training with more than two iterations seems to simply cause training to become less stable and introduces slight performance degradation.

\begin{figure}
\begin{center}
\resizebox{\linewidth}{!}{ %
\begin{tabular}{l|ccc|c}
\toprule
loss & w/o $\loss{poly}$  & w/o $\loss{shift}$ & w/o $\loss{sem}$ & Full    \\
\midrule
mAP & {$95.42_{\pm 0.79}$} & {${95.38}_{\pm 0.34}$} & $95.06_{\pm 0.62}$ & $\textbf{96.80}_{\pm 0.35}$
\\  
\bottomrule
\end{tabular}}
\includegraphics[width=\linewidth,trim=0 0 0 -4em]{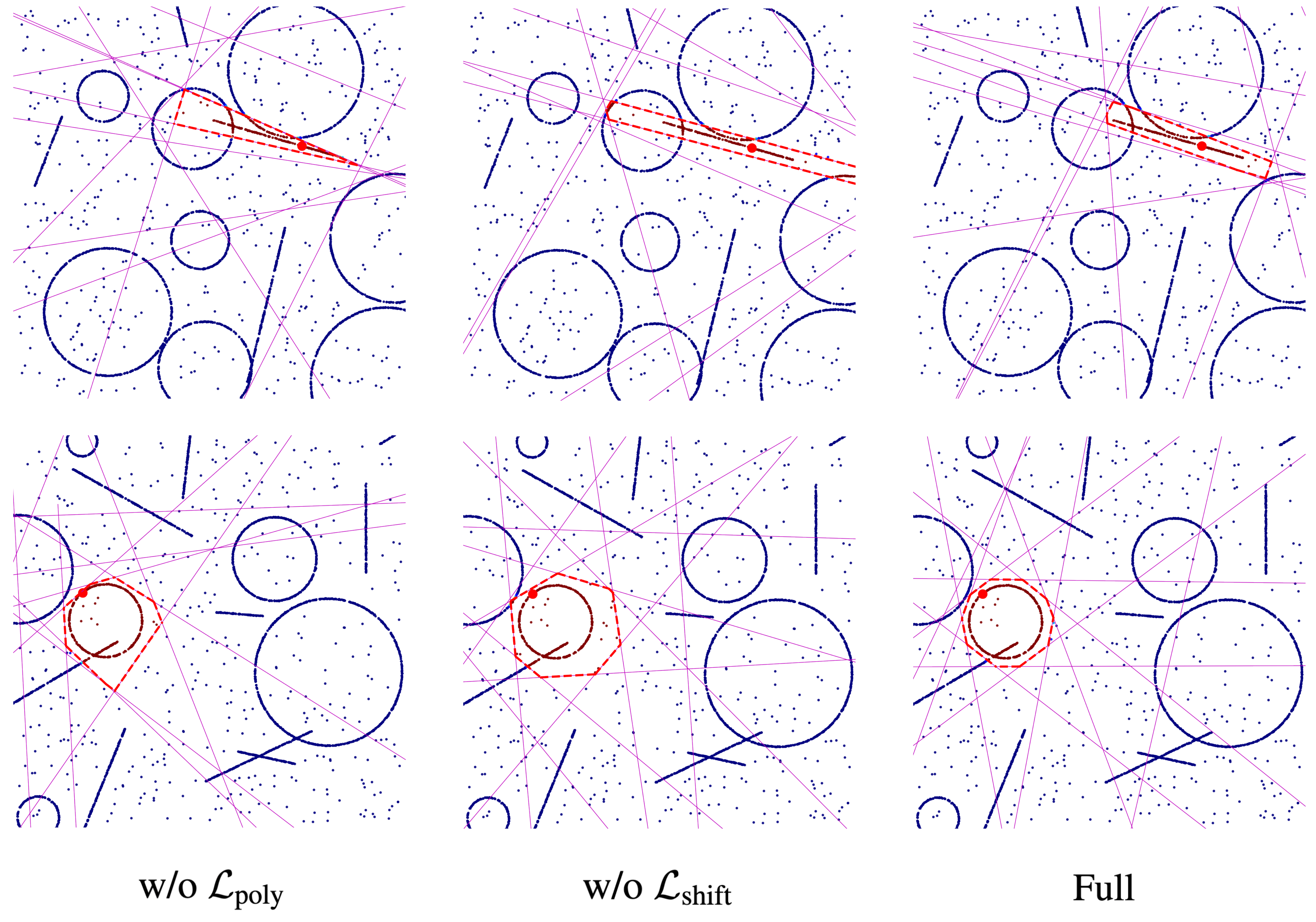}
\end{center}
\vspace{-1em}
\caption{
\textbf{Ablation: losses --}
We report the average and standard deviation of mAP by repeating the experiment 5 times.
} %
\label{fig:abs_loss}
\end{figure}
\paragraph{Losses -- \Fig{abs_loss}}
With the proposed regularizers, our algorithm learns tighter instance polytopes (w/ $\loss{poly}$ and $\loss{offset}$) and semantic similarity (w/ $\loss{sem}$), leading to significantly improved performance. 
Note that, since we evaluate AP for each the semantic category, we provide ground-truth semantic label for the models trained without semantic prediction (i.e., w/o $\loss{sem}$), 
Finally, note also that even without $\loss{poly}$ and $\loss{offset}$, our algorithm can still learn polytopes that \textit{roughly} segment instances.
\section{Conclusions}
We have proposed an instance proposal method for point clouds.
We formulate instance proposals as a query-conditioned attention model and employ neural bilateral filtering to provide much more accurate proposals than direct regression.
We demonstrate through synthetic data that the proposal generation process is indeed a bottleneck, which our method can significantly improve.
We further demonstrate the potential of our method on the ScanNet dataset, achieving competitive performance amongst top-down methods.

\paragraph{Limitations and Future works}
While we have shown clearly that a bottleneck exists, and that it can be avoided, its benefit has not been as strikingly revealed when compared to pipelines that are carefully engineered for instance segmentation.
We believe there is much room for this potential to be realized, similar to how top-down methods are the dominant strategy for natural images~\cite{wang2021solo,wang2020solov2}.
\section*{Acknowledgements}
This work was supported by the Natural Sciences and Engineering Research Council of Canada (NSERC) Discovery Grant, NSERC Collaborative Research and Development Grant, Google, Digital Research Alliance of Canada, and Advanced Research Computing at the University of British Columbia.
{
    \small
    \bibliographystyle{ieee_fullname}
    \bibliography{macros,main}
}

\appendix

\twocolumn[
\centering
\Large
\textbf{NeuralBF: Neural Bilateral Filtering \\ for Top-down Instance Segmentation on Point Clouds} \\
\vspace{0.5em}Supplementary Material \\
\vspace{1.0em}
] %
\appendix
\setcounter{page}{1}
\begin{figure*}[h]
\begin{center}
\includegraphics[width=.94\linewidth]{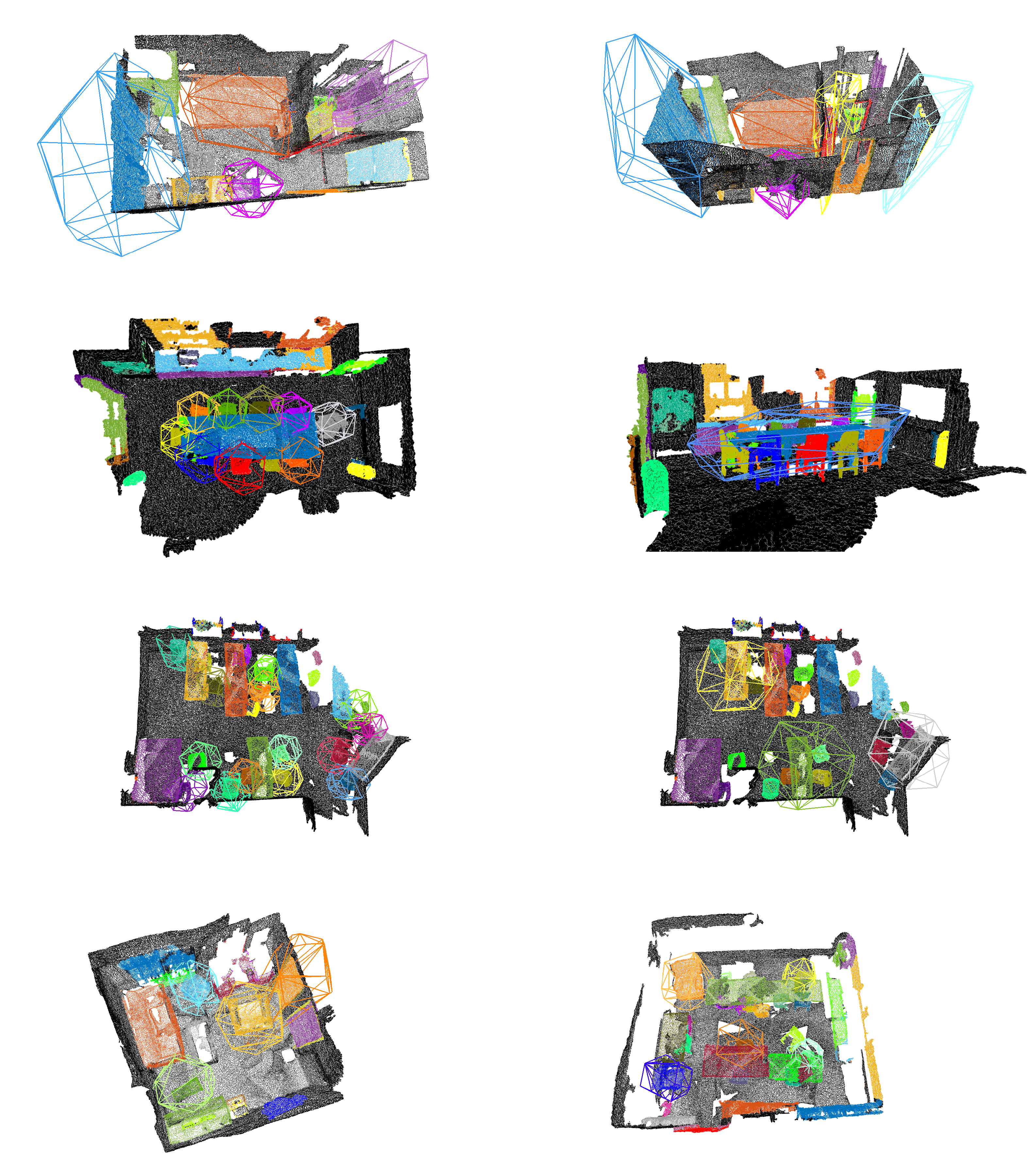}
\end{center}
\vspace{-1em}
\caption{
\textbf{Visualization of convex hulls on ScanNet -- } Note we visualize only some of the hulls to reduce visual density. We visualize the wireframe of the hulls for better visibility. And we color the points according to the instance prediction. 
}
\label{fig:cvx_hulls_3d}
\end{figure*}

\section{Visualization of convex hulls in ScanNet -- \Fig{cvx_hulls_3d}}
We further illustrate the learned convex hulls in ScanNet. 
As shown in \Fig{cvx_hulls_3d}, the learned convex hulls act as the rough bounding box of the queried instance. 

\begin{figure*}[h]
\begin{center}
\includegraphics[width=.90\linewidth]{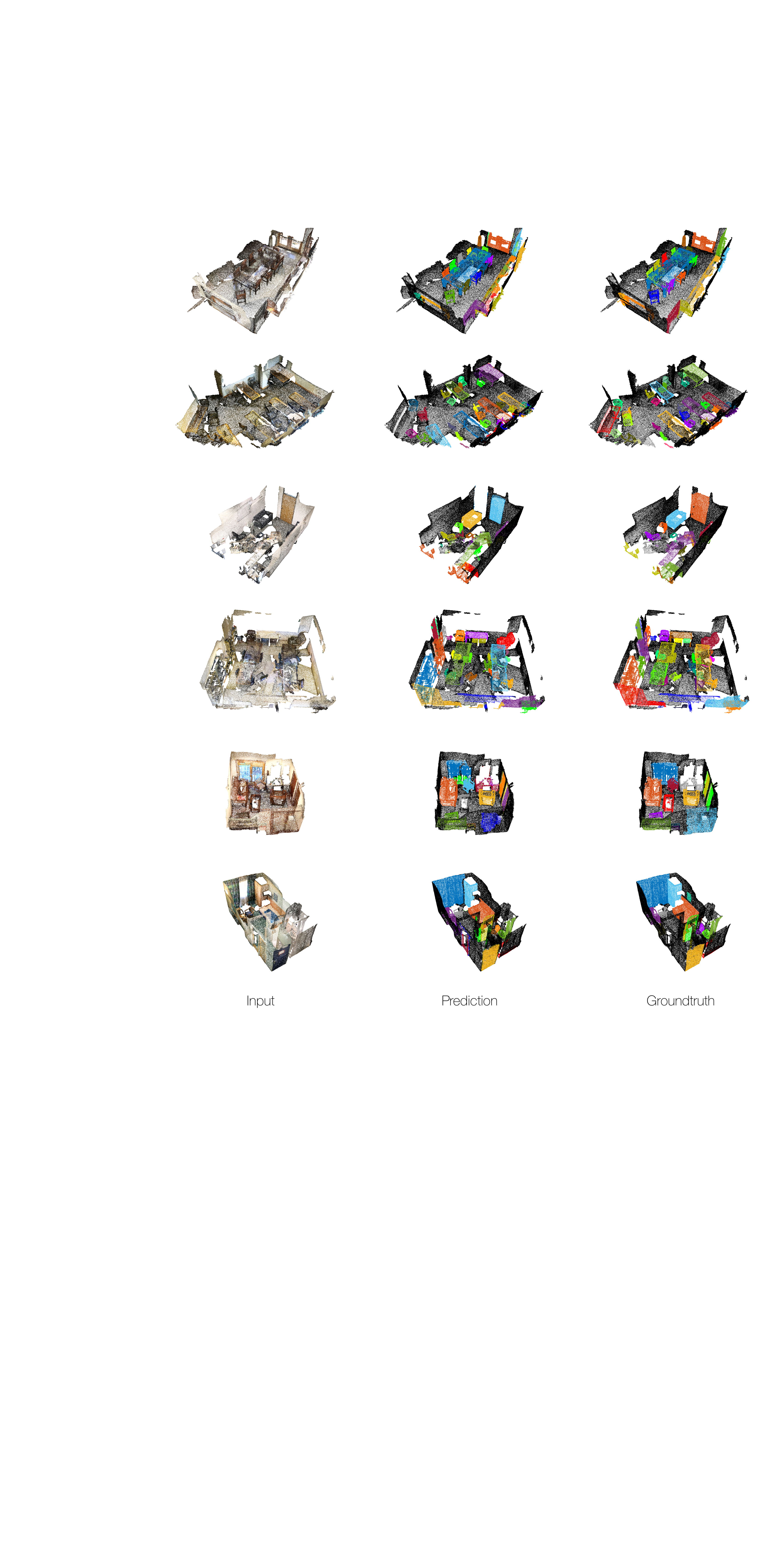}
\end{center}
\vspace{-1em}
\caption{
\textbf{More qualitative results on ScanNet --}
Instance segmentation results in the validation set.
}
\label{fig:scannet_supp_quali}
\end{figure*}
\section{More qualitative results in ScanNet -- \Fig{scannet_supp_quali}}
In addition to qualitative results in the main paper, we provide more qualitative results in Scannet.

\section{Interactive 3D visualization}
We further provide the interactive visualization of convex hulls and instance segmentation. 
It's available at: \url{https://nbf-supplementary.github.io}.

\end{document}